\documentclass{article}
\PassOptionsToPackage{numbers,sort}{natbib}
\usepackage[dblblindworkshop, final]{neurips_2025}
\workshoptitle{Mechanistic Interpretability Workshop at NeurIPS 2025}
\usepackage[utf8]{inputenc} 
\usepackage[T1]{fontenc}    
\usepackage[colorlinks = true,
            linkcolor = black,
            urlcolor  = blue,
            citecolor = blue,
            anchorcolor = blue]{hyperref}
\usepackage{url}            
\usepackage{booktabs}       
\usepackage{amsfonts}       
\usepackage{nicefrac}       
\usepackage{microtype}      
\usepackage[dvipsnames]{xcolor}
\usepackage{amsmath}
\usepackage{booktabs}

\usepackage{tabularx}
\usepackage{comment}
\usepackage{graphicx}
\usepackage[most]{tcolorbox}
\usepackage{subcaption}
\usepackage{caption}

\definecolor{myPurple}{RGB}{128,0,128} 

\title{RelP: Faithful and Efficient Circuit Discovery in Language Models via Relevance Patching}

\author{
Farnoush Rezaei Jafari$^{1,2}$\thanks{Correspondence to: \href{mailto:rezaeijafari@campus.tu-berlin.de}{rezaeijafari@campus.tu-berlin.de}}\\
\And 
Oliver Eberle$^{1,2}$ \\
\And
Ashkan Khakzar \\
\And
Neel Nanda 
\\ \\
\footnotesize{$^{1}$Machine Learning Group, Technische Universit\"at Berlin}\\
\footnotesize{$^{2}$BIFOLD -- Berlin Institute for the Foundations of Learning and Data}\\
}

\begin{document}
\maketitle

\begin{abstract}
Activation patching is a standard method in mechanistic interpretability for localizing the components of a model responsible for specific behaviors, but it is computationally expensive to apply at scale. Attribution patching offers a faster, gradient-based approximation, yet suffers from noise and reduced reliability in deep, highly non-linear networks.
In this work, we introduce \emph{Relevance Patching} (RelP), which replaces the local gradients in attribution patching with propagation coefficients derived from Layer-wise Relevance Propagation (LRP). LRP propagates the network's output backward through the layers, redistributing relevance to lower-level components according to local propagation rules that ensure properties such as relevance conservation or improved signal-to-noise ratio. Like attribution patching, RelP requires only two forward passes and one backward pass, maintaining computational efficiency while improving faithfulness.
We validate RelP across a range of models and tasks, showing that it more accurately approximates activation patching than standard attribution patching, particularly when analyzing residual stream and MLP outputs in the Indirect Object Identification (IOI) task. For instance, for MLP outputs in GPT-2 Large, attribution patching achieves a Pearson correlation of 0.006, whereas RelP reaches 0.956, highlighting the improvement offered by RelP.
Additionally, we compare the faithfulness of sparse feature circuits identified by RelP and Integrated Gradients (IG), showing that RelP achieves comparable faithfulness without the extra computational cost associated with IG. Code is available at \url{https://github.com/FarnoushRJ/RelP.git}.

\end{abstract}

\section{Introduction}
Recent advances in machine learning continue to rely on transformer-based language models, which achieve remarkable performance across a wide range of tasks. Given their widespread adoption, understanding the internal mechanisms of these models is an important challenge for improving our ability to interpret, trust, and control them. 

To address this challenge, the fields of eXplainable Artificial Intelligence (XAI) \cite{xaireview2021, longo2024explainable} and, more recently, mechanistic interpretability \cite{olah2020zoom, sharkey2025openproblemsmechanisticinterpretability} have emerged, aiming to reveal the decision-making processes of such complex architectures. Early efforts in XAI focused on developing feature attribution methods, often visualized as heatmaps, to highlight relevant features in classification models. As state-of-the-art models have grown more complex, approaches that extend beyond input explanations have become increasingly important for uncovering their internal strategies. This has motivated a shift toward exploring the complex internal structure of neural networks through methods such as concept-based interpretability \cite{chormai2024disentangled, prism2025}, higher-order explanations \cite{Eberle2022, schnake2022higher, symbXAI}, and circuit discovery \cite{olah2020zoom, wang2022interpretability}. Herein, a central goal of interpretability research is the reliable localization of specific model components responsible for particular functions, algorithms or behaviors \cite{olah2020zoom, zhong2023clock, sharkey2025openproblemsmechanisticinterpretability, ito2025quantifying, algeval2025}. To identify components that are causally involved in inference, mechanistic interpretability has proposed activation patching, also known as causal mediation analysis \cite{chan2022causal, geiger2022inducing, meng2022locating}. Activation patching replaces the activations of selected components in an original input with those from a patch input, allowing direct causal testing of the contribution of a component to the model's output.

As these patching-based interventions are computationally expensive, it is challenging to scale them, especially for larger models. To improve efficiency, attribution patching \cite{AtP} was introduced as a fast gradient-based approximation to activation patching. Using gradients to estimate the influence of each component on the model's output, attribution patching significantly reduces the computational cost compared to traditional perturbation-based approaches.  
However, this efficiency comes at the expense of robustness and accuracy, particularly in deep models like state-of-the-art LLMs \cite{kramár2024atpefficientscalablemethod}.
This issue is not unique to attribution patching. Many gradient-based attribution methods \cite{smilkov2017smoothgrad, gradcam, intgrad, GI, baehrens10a} suffer from a lack of robustness in large networks, often due to noisy and unreliable gradients \cite{balduzzi2017shattered} that undermine the quality of explanations \cite{DOMBROWSKI2022108194, DBLP:conf/icml/AliSEMMW22, pmlr-v235-achtibat24a}.

Alternative attribution methods have been proposed to overcome these limitations. One of the most widely used methods is Layer-wise Relevance Propagation (LRP) \cite{bach2015pixel}, developed within the theoretical framework of Deep Taylor Decomposition (DTD) \cite{montavon2017explaining}. LRP works by propagating the model's output backward through the network, redistributing the relevance of each component to its inputs according to layer-specific propagation rules. These rules are designed to enforce desirable properties such as conservation of total relevance, sparsity of explanations, and robustness to noise.

Building on these insights, we propose \emph{Relevance Patching} (RelP), a technique that replaces local gradients in attribution patching with propagation coefficients achieved using LRP. RelP enables more faithful localization of influential components in large models, without sacrificing scalability. Similar to attribution patching, RelP requires two forward passes and one backward pass.
We demonstrate the effectiveness of RelP in the context of the Indirect Object Identification task, showing that it consistently outperforms attribution patching across a range of architectures and model sizes, including GPT-2 \{Small, Medium, Large\} \cite{gpt2}, Pythia-\{70M, 410M\} \cite{pythia}, Qwen2-\{0.5B, 7B\} \cite{qwen}, and Gemma2-2B \cite{gemma}, with the strongest gains on the residual stream and MLP outputs. In addition, we apply RelP to recover sparse feature circuits responsible for Subject–Verb Agreement in the Pythia-70M model. Compared to Integrated Gradients \cite{intgrad}, RelP demonstrates comparable faithfulness in identifying meaningful circuits while also offering greater computational efficiency.

\begin{figure}[t]
    \centering
    \includegraphics[width=\linewidth]{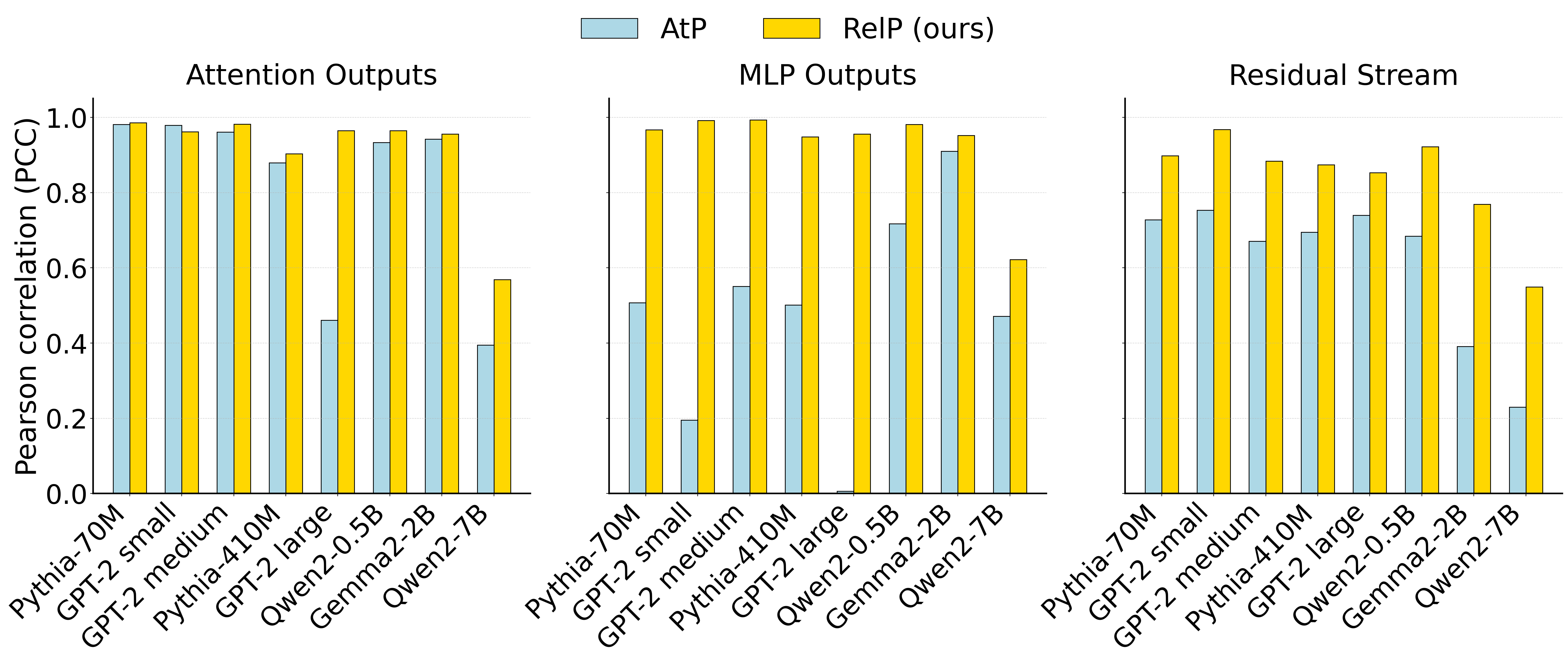}
    \caption{Pearson correlation coefficient (PCC) between activation patching and attribution patching (AtP) or relevance patching (RelP), computed over 100 IOI prompts for three GPT-2 model sizes (Small, Medium, Large), two Pythia models (70M, 410M), two Qwen2 models (0.5B, 7B), and Gemma2-2B. A higher value of PCC represents higher alignment with activation patching results.}
    \label{fig:quant_results}
\end{figure}

\section{Related Works}
\paragraph{Activation and Attribution Patching Techniques}
Activation patching is a causal mediation technique that tests the role of model components by replacing activations from an original run with those from a patch run \cite{pearl2000causality, vig2020investigating}. It has been widely used in interpretability studies \cite{geiger2022inducing, soulos-etal-2020-discovering, geiger2020neural, finlayson-etal-2021-causal}, with variants such as causal tracing, which perturbs activations with Gaussian noise \cite{meng2022locating}, and path patching, which extends interventions to multi-step computation paths \cite{goldowskydill2023localizing, wang2019structured}. To improve efficiency, attribution patching (AtP) uses gradient-based attribution instead of full interventions, requiring only two forward passes and one backward pass \cite{AtP}, and AtP* further enhances reliability while maintaining scalability \cite{kramár2024atpefficientscalablemethod}. Building on this line of work, we propose Relevance Patching (RelP), which also targets efficiency but replaces the local gradients in attribution patching with propagation coefficients computed using Layer-wise Relevance Propagation (LRP) \cite{bach2015pixel}, improving faithfulness to activation patching while maintaining the computational advantages of AtP.

\paragraph{Circuit Analysis}
Circuit analysis methods aim to uncover subnetworks that drive particular model behaviors. Activation patching, used in ACDC \cite{conmy2023towards}, identifies critical edges in the computation graph but is computationally expensive. It has been scaled to large models \cite{lieberum2023doescircuitanalysisinterpretability} and adapted to study the effects of fine-tuning \cite{prakash2024finetuning}. Sparse Autoencoders (SAEs) offer an alternative by learning disentangled, interpretable latent features \cite{marks2024sparse, kharlapenko2025scaling}. Methods such IFR \cite{ferrando2024information} and EAP \cite{syed2023attribution} instead use gradient-based attributions to perform circuit discovery in a more efficient way. Finally, attribution-guided pruning with LRP has been shown to support both circuit discovery and model compression \cite{hatefi2025attributionpruning}. Our proposed Relevance Patching (RelP) also builds on LRP but stays within the patching framework, providing faithful and efficient attribution that approximates activation patching.

\paragraph{Gradient-Based and Propagation-Based Attribution Methods}
Local gradients have long been used to explain the behavior of non-linear models \cite{baehrens10a}. Through backpropagation, they enable the efficient computation of saliency maps \cite{bach2015pixel, saliency}, which have been extensively studied in the context of vision models. Saliency maps derived from raw gradients \cite{saliency} capture the sensitivity of model predictions to small perturbations in input features. When these gradients are scaled by the corresponding input values, they yield Gradient$\times$Input \cite{GI} explanations. Grad-CAM \cite{gradcam} aggregates class-specific gradients over the final convolutional feature maps, yielding coarse but discriminative heatmaps. Since raw gradients are often noisy, SmoothGrad \cite{smilkov2017smoothgrad} averages saliency maps over random perturbations, PathwayGrad \cite{khakzar2021neural} finds a sub-network (pathway) critical for the output and propagates gradients through that pathway, and Integrated Gradients \cite{intgrad} integrates along a baseline-to-input path to mitigate saturation and enforce key attribution axioms. Beyond gradient-based methods, propagation-based approaches such as LRP \cite{bach2015pixel} and DeepLIFT \cite{shrikumar2017learning} redistribute the model’s output backward through the network using local propagation rules, which can offer more faithful attribution. For a comprehensive review of explanation methods and attribution techniques, see \cite{montavon2019layer, xaireview2021, khakzar2022explanations, longo2024explainable}.

\section{Background}
\label{sec:background}
\paragraph{From Activation to Attribution Patching}
Let $\mathcal{M}: X \rightarrow \mathbb{R}^V$ be a decoder-only transformer model that maps an input sequence $x \in X:= \{1, ..., V\}^T$ to a vector of output logits over a vocabulary of size $V$. We represent the model as a directed computation graph $G=(N, E)$, where each node $n \in N$ corresponds to a distinct model component, and each edge $(n_1, n_2) \in E$ represents a direct computational dependency. The activation of a component $n$ when processing input $x$ is denoted $n(x)$.

Let \( D \) be a distribution over prompt pairs \( (x_{\text{original}}, x_{\text{patch}}) \), where \( x_{\text{original}} \) is a representative task sample that captures the behavior of interest, and \( x_{\text{patch}} \) serves as a reference input used to introduce counterfactual perturbations for causal intervention. A metric $\mathcal{L}: \mathbb{R}^V \rightarrow \mathbb{R}$ quantifies the output behavior of the model.

The causal contribution $c(n)$ of a component $n \in N$ is defined as the expected effect of replacing its activation in the original input with the corresponding activation from the patch input, i.e.,

\begin{equation}
c(n) := \mathbb{E}_{(x_{\text{original}}, x_{\text{patch}}) \sim D} \left[ 
\mathcal{L}\left( \mathcal{M}(x_{\text{original}} \mid \text{do}(n \leftarrow n(x_{\text{patch}}))) \right)
- 
\mathcal{L}\left( \mathcal{M}(x_{\text{original}}) \right)
\right] .
\end{equation}

This definition follows the causal intervention framework, where $\text{do}(n \leftarrow n(x_{\text{patch}}))$ denotes overwriting the activation of $n$ with its value under the patch input. Evaluating $c(n)$ directly for all $n \in N$ is computationally prohibitive for large-scale models, motivating the development of efficient approximation methods such as attribution patching (AtP) \cite{AtP}: 

\begin{equation}
\hat{c}_{\text{AtP}}(n) := \mathbb{E}_{(x_{\text{original}}, x_{\text{patch}}) \sim D}
\left[
\left(n(x_{\text{patch}}) - n(x_{\text{original}}) \right)^\top
\left. \frac{\partial \mathcal{L}(\mathcal{M}(x_{\text{original}}))}{\partial n} \right|_{n = n(x_{\text{original}})}
\right].
\label{eq:AtP}
\end{equation}

AtP (Eq. \ref{eq:AtP}) approximates the effect of replacing component $n$'s activation from the original input with that from the patch input, without explicitly running the patched model, by leveraging gradients. 

\paragraph{Layer-wise Relevance Propagation (LRP)}
Layer-wise Relevance Propagation (LRP) \cite{bach2015pixel} is a framework for interpreting neural network predictions. The central idea is to assign a relevance score $\mathcal{R}$ to each component or input feature, quantifying its contribution to the model's output or to any metric defined from the output. These relevance scores are then propagated backward through the network according to layer-specific rules. The propagation follows the principle of relevance conservation \cite{bach2015pixel, montavon2017explaining}, ensuring that the total relevance at each layer is completely redistributed to the preceding layer, ultimately attributing the chosen metric to the input features.

To formalize this process, let $a_i^{(l-1)}$ with $i\in\{1,...,n\}$ denote the activations in layer $l-1$. These activations serve as inputs to a layer represented by $f^{(l)}:\mathbb{R}^n \rightarrow \mathbb{R}^m$, resulting in outputs $a_j^{(l)}$, $j\in\{1,...,m\}$. A local first-order Taylor expansion around a reference point $\tilde a^{(l-1)}$ expresses each output as
\[
f^{(l)}_j(a^{(l-1)}) \approx \sum_i J^{(l)}_{ji}\, a_i^{(l-1)} + \tilde b^{(l)}_j, 
\qquad 
J^{(l)}_{ji} = \left. \frac{\partial f^{(l)}_j}{\partial a_i^{(l-1)}} \right|_{\tilde a^{(l-1)}}.
\]
Here, $J^{(l)}_{ji}$ is the local Jacobian, describing how activations $a_i^{(l-1)}$ influences outputs $a_j^{(l)}$, while $\tilde b^{(l)}_j$ collects constant terms and approximation errors. 

This decomposition motivates the LRP redistribution step. Each component $j$ in layer $l$ is assigned a relevance score $\mathcal{R}^{(l)}_j$, which is redistributed to its inputs in proportion to their contributions to the activation of component $j$. Here, this redistribution from upper-layer activations to component $i$ is expressed through a \textbf{propagation coefficient} $\rho_i$, which aggregates the contributions from all connected upper-layer components $j$. The relevance of component $i$ in layer $l-1$ is then given by
\[
\mathcal{R}^{(l-1)}_i = a_i^{(l-1)}\, \rho_i.
\]

The propagation coefficient $\rho_i$ depends on the local Jacobian $J^{(l)}$, the activations $a^{(l-1)}$, and the relevance scores at layer $l$ (i.e., $\mathcal{R}^{(l)}$). Intuitively, $\rho_i$ acts as a filter that determines which parts of the input activations are considered relevant.

The layer-wise redistribution of relevance can be repeated until the input layer is reached or stopped at an intermediate layer of interest. The procedure begins with an initial relevance signal \(\mathcal{R}^{(L)}_c\), typically defined using a metric computed from the model outputs, commonly the logit of the true, predicted, or any target class \(c\) at the output layer \cite{bach2015pixel,montavon2017explaining}. Contrastive approaches have also been proposed, which explain the difference between class logits to highlight features that strongly influence changes in the model's prediction \cite{DBLP:conf/icml/AliSEMMW22, contrastivelrp23}.

\section{RelP: Relevance Patching for Mechanistic Analysis}
Gradient-based attribution methods have been extensively studied in the XAI literature \cite{baehrens10a, GI, smilkov2017smoothgrad, intgrad, gradcam}. Despite their simplicity and model-agnostic applicability, these methods often face challenges when applied to large neural networks due to noisy and less reliable gradients \cite{DOMBROWSKI2022108194, pmlr-v235-achtibat24a}. 
LRP mitigates these shortcomings by propagating the network's output $\mathcal{M}(x)$, or any metric $\mathcal{L}$ defined from the output, backward through the network in a structured and theoretically grounded manner. Depending on the layers in a given model architecture, different propagation rules have been proposed. These rules are typically derived from gradient analyses of model components \cite{montavon2017explaining, DBLP:conf/icml/AliSEMMW22, jafari2024mambalrp}, with specific choices guided by common desiderata for explanations, such as sparsity, reduced noise, and relevance conservation \cite{10.1007/978-3-642-77927-5_24,  montavon2019layer}. 

Specialized propagation rules have been developed to handle diverse nonlinear components (e.g., activation functions, attention mechanisms, normalization layers) and architectural nuances across deep network families \cite{Arras2019, jafari2024mambalrp, pmlr-v235-achtibat24a, montavon2019layer, Eberle2022, letzgus2022, schnake2022higher}. An overview of relevant rules for transformer models is shown in Table~\ref{tab:lrp_rules}.

As attribution patching relies on gradients to approximate the effect of substituting hidden activations (“patching in”), it is also susceptible to the noise and approximation errors inherent to gradient-based attribution methods, particularly in very large models \cite{balduzzi2017shattered, montavon2017explaining}. To address this, we introduce \emph{Relevance Patching} (RelP).  

Formally, RelP follows the structure of standard attribution patching (AtP). In AtP, the contribution of a component is computed as the dot product between the change in its activation, caused by replacing the original input with a patch input, and the gradient of the metric $\mathcal{L}$ with respect to that component, evaluated at the original input. RelP modifies this procedure by substituting the gradient term with the LRP-derived propagation coefficient (introduced in Section~\ref{sec:background}) for that component. The resulting contribution score is defined as

\begin{align}
\hat{c}_{\text{RelP}}(n) &:= 
\mathbb{E}_{(x_{\text{original}}, x_{\text{patch}}) \sim D} 
\Big[
\left( n(x_{\text{patch}}) - n(x_{\text{original}}) \right)^\top
\textcolor{violet}{\rho(\mathcal{L}(\mathcal{M}(x_{\text{original}})))} \Big|_{n(x_{\text{original}})}
\Big] \notag \\
&= \mathbb{E}_{(x_{\text{original}}, x_{\text{patch}}) \sim D} 
\Big[
\textcolor{violet}{\mathcal{R_{\text{RelP}}}(\mathcal{L}(\mathcal{M}(x_{\text{original}})))} \Big|_{n(x_{\text{original}})}
\Big],
\label{eq:LRP-P}
\end{align}

where $\rho(\mathcal{L}(\mathcal{M}(x_{\text{original}})))$ and $\mathcal{R}_{\text{RelP}}(\mathcal{L}(\mathcal{M}(x_{\text{original}})))$ denote, respectively, the propagation coefficient and the relevance score for component $n$.

RelP preserves the efficiency of attribution patching while improving faithfulness. Unlike \citet{marks2024sparse}, which relies on Integrated Gradients \cite{intgrad} for more accurate approximations at higher computational cost, RelP provides scalable and faithful localization of influential components. 

\paragraph{Propagation Rules}Table~\ref{tab:lrp_rules} summarizes the propagation rules applicable to core components of transformer architectures. As discussed in Section~\ref{sec:background}, relevance conservation is a key property in LRP, ensuring that the sum of relevance scores is preserved across layers. This prevents relevance from being lost or artificially introduced during propagation. Many of the specialized rules in Table~\ref{tab:lrp_rules} are designed to address situations where this property could fail, and are often derived from gradient analyses of individual model components.

\begin{table}
\begin{tcolorbox}[
  floatplacement=htp,
  colback=cyan!2!white,    
  colframe=cyan!50!white,    
  title=Propagation Rules for Transformers,
  fonttitle=\bfseries\small,
  coltitle=black,
  boxrule=1.5pt,
  enhanced
]
\centering
\small
 \resizebox{\textwidth}{!}{
\begin{tabularx}{\linewidth}{l l X}
\toprule
\textbf{Layer} & \textbf{Propagation Rule} & \textbf{Implementation Trick} \\
\midrule
LayerNorm / RMSNorm & LN-rule \cite{DBLP:conf/icml/AliSEMMW22} & \( y_i = \frac{x_i - E[x]}{\textcolor{myPurple}{[}\sqrt{\epsilon + \text{Var}[x]}\textcolor{myPurple}{]}_{\textcolor{myPurple}{\text{const.}}}} \) \\
GELU / SiLU & Identity-rule \cite{jafari2024mambalrp} & \( x \odot \textcolor{myPurple}{[}\Phi(x)\textcolor{myPurple}{]}_{\textcolor{myPurple}{\text{const.}}} \) \\
Linear & 0-rule, \(\epsilon\)-rule, or \(\gamma\)-rule \cite{montavon2019layer} & -- \\
Attention & AH-rule \cite{DBLP:conf/icml/AliSEMMW22} & \( y_j = \sum_i x_i \textcolor{myPurple}{[}A_{ij}\textcolor{myPurple}{]}_{\textcolor{myPurple}{\text{const.}}} \) \\
Multiplicative Gate & Half-rule \cite{jafari2024mambalrp, Arras2019} & \( 0.5 \cdot (x \odot g(x)) + 0.5 \cdot \textcolor{myPurple}{[}(x \odot g(x))\textcolor{myPurple}{]}_{\textcolor{myPurple}{\text{const.}}} \) \\
\bottomrule
\end{tabularx}
}
\vspace{0.7em}
\captionof{table}{A summary of propagation rules for Transformer layers. Components marked with 
\textcolor{myPurple}{$\text{[]}_\text{const.}$} are treated as constants, typically using 
\textcolor{myPurple}{\texttt{.detach()}} in PyTorch. In this table, $x$ and $y$ denote the input 
and output of a layer, respectively, and $A$ represents the attention weights.}
\label{tab:lrp_rules}
\end{tcolorbox}
\end{table}

In attention heads, attention outputs depend on inputs directly and via attention weights $A_{ij}$, which are also input-dependent. Correlations between these terms can break conservation and the AH‑rule preserves it by treating $A_{ij}$ as constants, effectively linearizing the heads \cite{DBLP:conf/icml/AliSEMMW22}. LayerNorm’s centering and variance-based scaling can cause “relevance collapse”, which the LN‑rule mitigates this by treating $(\sqrt{\epsilon+\mathrm{Var}[x]})^{-1}$ as constant, allowing the operation to be seen as linear and preserving conservation \cite{DBLP:conf/icml/AliSEMMW22}. For multiplicative gates, the Half‑rule splits relevance equally to avoid spurious doubling \cite{jafari2024mambalrp, Arras2019}.
Activation functions (e.g., GELU, SiLU) can disrupt conservation since their output is a nonlinear transformation of the input, the Identity‑rule addresses this by treating the non‑linear component as constant \cite{jafari2024mambalrp}. For linear layers, the $0$‑rule is equivalent to Gradient $\times$Input~\cite{GI}. The $\epsilon$‑rule and $\gamma$‑rule extend the $0$‑rule, the $\epsilon$‑rule is applied when sparsity and noise reduction are desired, while the $\gamma$‑rule is used when it is preferable to emphasize positive contributions over negative ones \cite{montavon2019layer}. For additional rules specifically tailored to transformers, we refer the readers to \cite{DBLP:conf/icml/AliSEMMW22, pmlr-v235-achtibat24a, bakish2025revisitinglrppositionalattribution}.

\section{Experiments}
\label{sec:ioi}
We evaluate the effectiveness of RelP by benchmarking it against attribution patching (AtP), on the Indirect Object Identification (IOI) task, across three GPT‑2 variants (Small, Medium, and Large), two Pythia models (70M and 410M), two Qwen2 models (0.5B and 7B), and Gemma2-2B. Furthermore, we assess RelP's ability to recover sparse feature circuits underlying Subject–Verb Agreement in the Pythia-70M model, comparing its performance against Integrated Gradients (IG) as a more accurate gradient-based baseline. Additional details of our experimental setup are provided in Appendix~\ref{sec:experimental_details}.

\subsection{Evaluating RelP on the IOI Task}
 The objective of this experiment is to evaluate how well attribution patching (AtP) and relevance patching (RelP) approximate the ground-truth effects captured by activation patching (AP) on the IOI task. This task involves determining which entity in a sentence functions as the indirect object, typically the recipient or beneficiary of the direct object.

To enable controlled evaluations, we construct 100 prompt pairs. Each pair consists of an original prompt (with the correct indirect object) and a patch variant in which the subject and indirect object are swapped, keeping all other tokens the same. All prompts are generated from structured templates \cite{wang2022interpretability} (Table~\ref{tab:ioi_templates}) to ensure consistency across examples. We define the metric $\mathcal{L}$ as the difference in logits between the original and patch targets. The methods RelP, AtP, and AP are each applied to the residual stream, attention outputs, and MLP outputs. Since the resulting attribution scores are computed across hidden dimensions, we aggregate them by summing over these dimensions.  To quantify how well AtP and RelP approximate AP, we compute the Pearson correlation coefficient (PCC) between their results and those of AP.
\begin{table}
\begin{tcolorbox}[
  floatplacement=htp,
  colback=green!5!white,    
  colframe=green!70!red,    
  title=IOI Templates,
  fonttitle=\bfseries\small,
  coltitle=black,
  boxrule=1.5pt,
  enhanced
]
\small
\centering
\begin{tabular}{c|l}
\toprule
\textbf{Template \#} & \textbf{Sentence Template} \\
\midrule
1 & Then, [B] and [A] went to the [PLACE]. [B] gave a [OBJECT] to [A]. \\
2 & When, [B] and [A] went to the [PLACE]. [B] gave a [OBJECT] to [A]. \\
3 & After [B] and [A] went to the [PLACE], [B] gave a [OBJECT] to [A]. \\
\bottomrule
\end{tabular}
\vspace{0.7em}
\captionof{table}{Templates used for generating the IOI samples.}
\label{tab:ioi_templates}
\end{tcolorbox}
\end{table}

The results of this experiment are presented in Figure~\ref{fig:quant_results}. As shown, RelP consistently outperforms AtP in a range of architectures and model sizes, including GPT-2 \{Small, Medium, Large\}, Pythia-\{70M, 410M\}, Qwen2-\{0.5B, 7B\}, and Gemma2-2B. The performance gap is especially pronounced in the MLP output and residual stream analyses. Detailed numerical results are listed in Table~\ref{tab:pcc} in Appendix~\ref{sec:quantitative_results}.

\begin{figure}
    \centering
    \includegraphics[width=\linewidth]{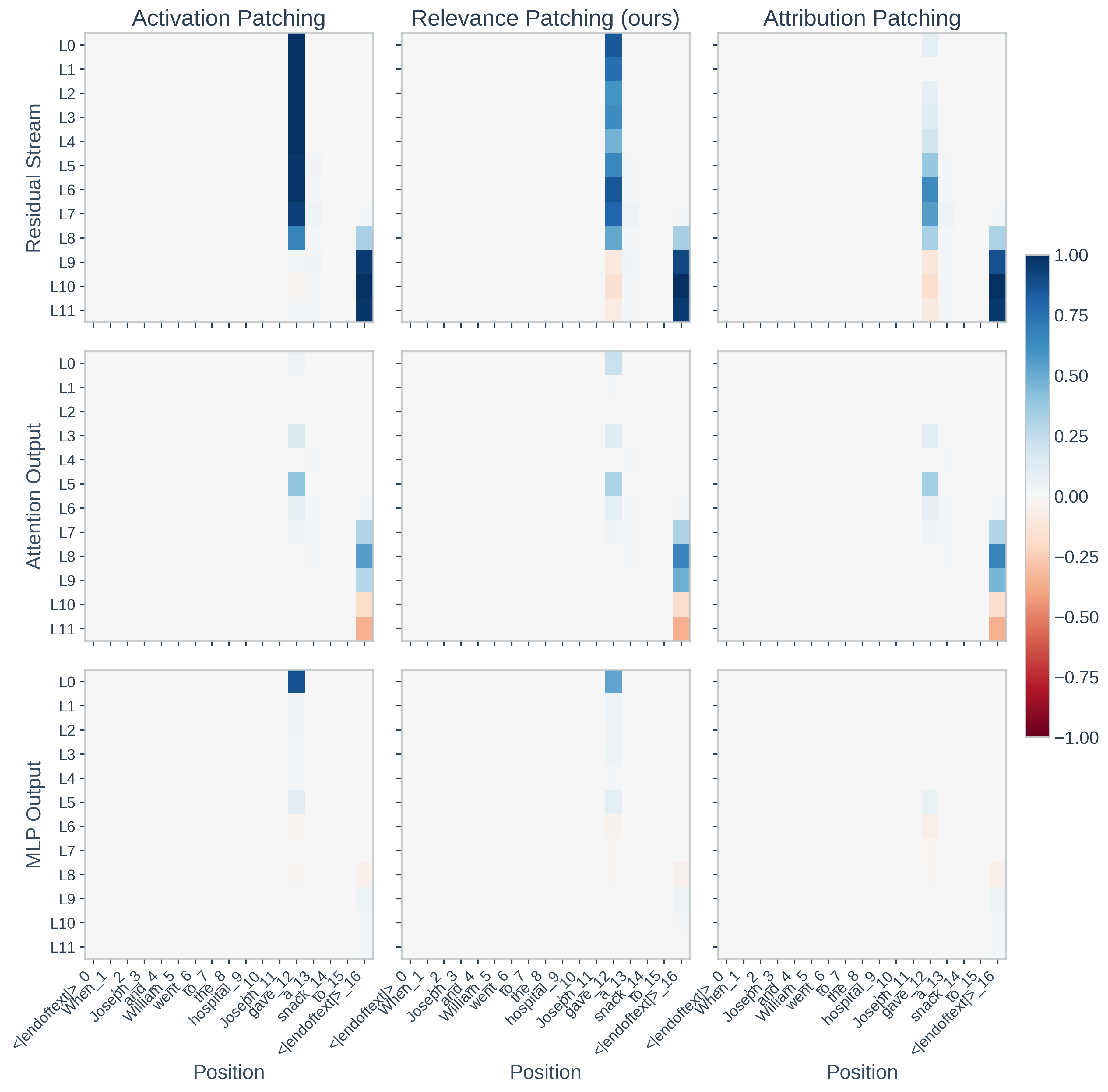}
    \caption{Qualitative comparison showing how accurately relevance patching (RelP) and attribution patching (AtP) approximate the effects of activation patching in GPT-2 Small. RelP shows notably better alignment in the residual stream and at MLP0, where AtP's estimates are less reliable.}
    \label{fig:ioi_qualitative}
\end{figure}

Qualitative differences between AtP and RelP are also visible in Figure~\ref{fig:ioi_qualitative}. \citet{AtP} suggests that attribution patching fails notably in the residual stream, primarily due to its large activations and the nonlinearity introduced by LayerNorm, which significantly disrupts the underlying linear approximation. It also performs poorly on MLP0, since this layer in GPT-2 Small functions as an ``extended embedding'', and \citet{AtP} has shown that linear approximations in this layer may become unstable, leading to misleading insights.
As can be seen in Figure~\ref{fig:ioi_qualitative}, the RelP results align more closely with those from activation patching, especially when considering the residual stream and MLP outputs (e.g., MLP0). Further qualitative results are provided in Appendix~\ref{sec:further_qualitative_results}.

Overall, we find that the proposed method, RelP, consistently achieves strong correlations with activation patching and is often clearly superior to standard attribution patching. This underscores the effectiveness of the tailored propagation rules in the LRP framework, which provide a reliable and computationally efficient approximation of the computationally expensive activation patching method across models and components commonly studied in mechanistic interpretability research.

\begin{figure}[htb]
    \centering
    \includegraphics[width=\linewidth]{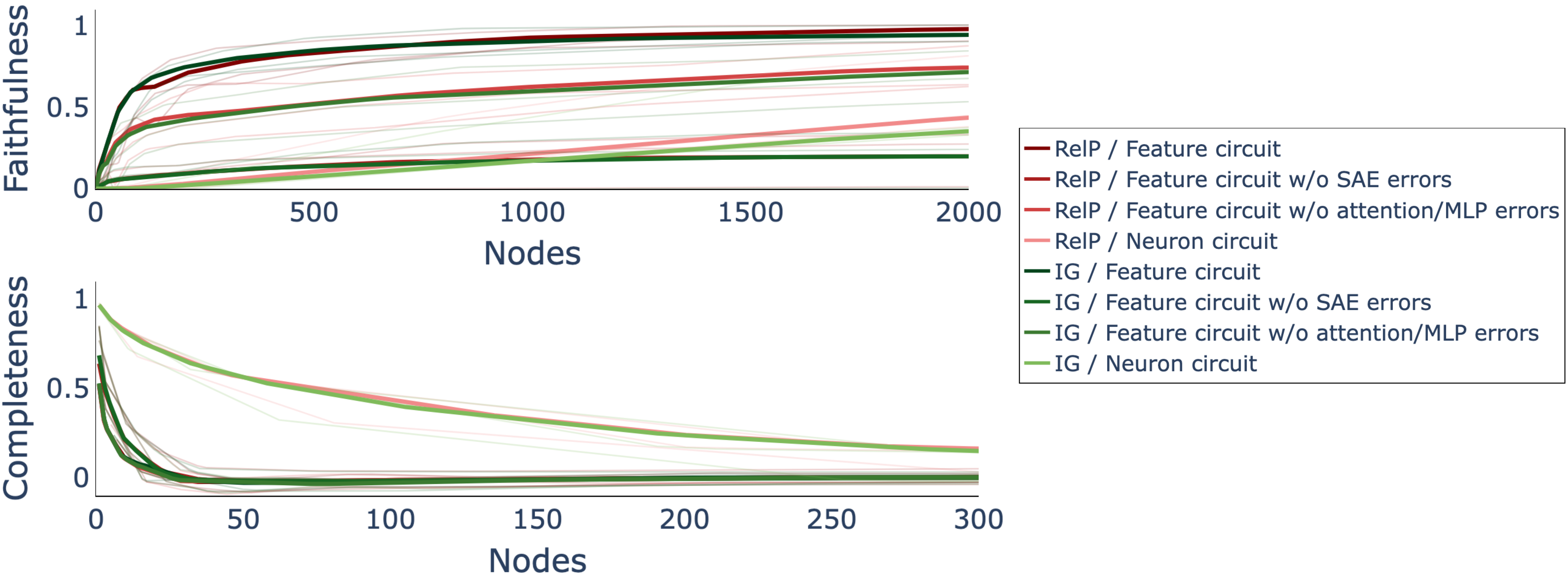}
    \caption{Faithfulness and completeness scores for circuits, evaluated on held-out data. Faint lines show individual circuits for structures from Table~\ref{tab:other_templates}, while the bold lines indicate the average across all structures. An ideal circuit has a faithfulness score of 1 and a completeness score of 0. While Integrated Gradients (IG) requires multiple integration steps (steps=10 in this experiment), RelP achieves comparable faithfulness scores without any additional computational cost.}
    \label{fig:faithfulness_completeness}
\end{figure}

\subsection{Sparse Feature Circuits for Subject-Verb Agreement}

Sparse feature circuits are small, interconnected groups of interpretable features that jointly drive specific behaviors in language models. Analyzing these circuits allows us to understand how models combine meaningful components to solve tasks, offering insight into the underlying mechanisms behind their decisions. In this experiment, we aim to uncover feature circuits involved in the subject-verb agreement task, a linguistic evaluation that tests a model’s ability to correctly match a verb’s inflection (e.g., singular or plural) to the grammatical number of its subject.

To identify and analyze these circuits, we use features discovered using sparse autoencoders (SAEs), which provide fine-grained, human-interpretable units. We follow \cite{marks2024sparse} and use their SAEs trained on Pythia-70M language model and their sparse features in our experiments.  These sparse features serve as the fundamental units for circuit analysis. To assess their influence on the model's output, \citet{marks2024sparse} employed efficient linear approximations of activation patching, such as Integrated Gradients (IG), to estimate the contributions of individual features and their interactions. Although IG provides greater faithfulness than attribution patching, it requires multiple forward and backward passes, resulting in higher computational cost. Our experiments show that RelP achieves faithfulness on par with attribution patching, without incurring additional computational overhead.

We use three templatic datasets (Table~\ref{tab:other_templates}), where tokens at the same positions serve similar roles, allowing us to average node and edge effects across examples while retaining positional information. Consistent with \cite{marks2024sparse}, we assess the identified sparse feature circuits using \textbf{faithfulness} and \textbf{completeness} evaluation metrics \cite{wang2022interpretability}. Additionally, we compare our sparse feature circuits with neuron circuits, constructed by applying the same discovery method directly to individual neurons rather than to SAE features, serving as a baseline for evaluation.

\begin{table}
\begin{tcolorbox}[
  floatplacement=htp,
  colback=yellow!10!white,    
  colframe=yellow!80!red,    
  title=Subject-Verb Agreement Templates,
  fonttitle=\bfseries\small,
  coltitle=black,
  boxrule=1.5pt,
  enhanced
]
\small
\resizebox{\textwidth}{!}{
\begin{tabular}{>{\bfseries}p{3.8cm} p{5cm} p{5cm}}
\toprule
\textbf{Template Type} & \textbf{Example $x_{\text{original}}$} & \textbf{Example Output Metric} \\
\midrule
Within Relative Clause (RC) & The athlete that the \textcolor{blue}{managers} & $p(\textcolor{red}{\text{likes}}) - p(\textcolor{blue}{\text{like}})$ \\
Across Relative Clause (RC) & The \textcolor{blue}{athlete} that the managers like & $p(\textcolor{red}{\text{do}}) - p(\textcolor{blue}{\text{does}})$ \\
Across Prepositional Phrase (PP) & The \textcolor{blue}{secretaries} near the cars & $p(\textcolor{red}{\text{has}}) - p(\textcolor{blue}{\text{have}})$ \\
\bottomrule
\end{tabular}
}
\vspace{0.7em}
\captionof{table}{These templates involve contrastive pairs of inputs that differ only in the grammatical number of the subject, with the model's task being to choose the appropriate verb inflection.}
\label{tab:other_templates}
\end{tcolorbox}
\end{table}

\textbf{Faithfulness} measures how much of the model’s original performance is captured by the discovered circuit, relative to a baseline where the entire model is mean-ablated. It is calculated as $\frac{\mathcal{L}(C) - \mathcal{L}(\emptyset)}{\mathcal{L}(M) - \mathcal{L}(\emptyset)}$, where $\mathcal{L}(C)$ is the metric when only the circuit $C$ is active, $\mathcal{L}(\emptyset)$ is the metric of the fully mean-ablated model, and $\mathcal{L}(\mathcal{M})$ is the metric of the full model.

\textbf{Completeness} evaluates how much of the model’s behavior is not explained by the discovered circuit. It is defined as the faithfulness of the circuit’s complement, $(\mathcal{M} \backslash C)$. In other words, if the complement of the circuit still explains a lot of the behavior, then the original circuit is not complete in capturing the mechanism.

We report faithfulness scores for feature and neuron circuits as the node threshold $T_N$ is varied (Figure~\ref{fig:faithfulness_completeness}). The threshold keeps only nodes with contribution scores above $T_N$, retaining the most relevant ones. Consistent with \cite{marks2024sparse}, small feature circuits explain most of the model’s behavior: in Pythia-70M, about 100 features account for the majority of performance, whereas roughly 1,500 neurons are needed to explain half. 

SAE error nodes summarize reconstruction errors in the SAE, making them fundamentally different from single neurons. To analyze their impact, we evaluate faithfulness after removing all SAE error nodes, as well as those originating from attention and MLP modules. Consistent with findings by \citet{marks2024sparse}, we observe that removing residual stream SAE error nodes severely disrupts model performance, whereas removing MLP and attention error nodes has a less pronounced effect.

Notably, the circuits identified by our RelP method achieve faithfulness scores comparable to those obtained with IG, while offering greater computational efficiency. In our experiments, IG requires 10 integration steps, increasing computational overhead. In contrast, RelP matches the efficiency of AtP, requiring only two forward passes and one backward pass, while achieving faithfulness scores on par with IG. In the case of neuron circuits and feature circuits without attention/MLP errors, we observe an improvement in the faithfulness of RelP compared to IG.
This combination of faithfulness and efficiency makes RelP a more effective and practical approach for uncovering sparse feature circuits in large language models.

Following \cite{marks2024sparse}, we also evaluate completeness (Figure~\ref{fig:faithfulness_completeness}) and observe that ablating only a few nodes from feature circuits can drastically reduce model performance, whereas hundreds of neurons are needed for a similar effect for Pythia. This highlights the efficiency of the identified sparse feature circuits in capturing critical model behavior, as well as the usefulness of RelP in locating them.

\section{Conclusion}
In this paper, we introduced Relevance Patching (RelP) as an enhancement over the standard attribution patching method. RelP preserves the overall algorithmic structure of attribution patching while replacing its local gradient signal with the propagation coefficient derived from Layer-wise Relevance Propagation (LRP). This modification enables RelP to more accurately approximate the causal effects captured by activation patching, while remaining computationally efficient.

In our experiments on the Indirect Object Identification task, we demonstrated that RelP exhibits stronger alignment with activation patching across residual stream, attention, and MLP outputs compared to attribution patching. Furthermore, in comparing sparse feature circuits identified by RelP and Integrated Gradients (IG), we showed that RelP achieves comparable faithfulness without incurring the additional computational cost associated with IG. Our study thus demonstrates that methods developed for feature attribution can be effectively integrated into mechanistic interpretability, helping to advance our understanding of modern foundation models.

\section{Limitations and Future Work}
Applying RelP to localize relevant model components requires choosing appropriate rules within the LRP framework, which introduces some model-specific overhead compared to model-agnostic attribution methods. In our experiments, RelP consistently outperformed attribution patching on residual stream and MLP outputs, while gains for attention outputs were smaller. This suggests that a deeper analysis of gradient structure in self-attention, combined with more careful rule selection, could improve attribution faithfulness. Prior works have examined these challenges in the context of input-level feature attribution \cite{DBLP:conf/icml/AliSEMMW22, pmlr-v235-achtibat24a, jafari2024mambalrp}, but systematic studies of internal layers remain limited. For our experiments, we used small- and medium-scale open-access models, which allow full access to internals and gradients. The tasks were chosen to match standard circuit analysis benchmarks, relying on predefined original–patch input pairs and single-token prediction metrics. As a result, circuit discovery was limited to moderately complex behaviors. Extending RelP to more challenging settings, such as free-form text generation or scenarios without known counterfactual inputs, remains an important direction for future work.

\begin{ack}
We acknowledge support by the Federal Ministry of Research, Technology and Space (BMFTR) for BIFOLD
(ref. 01IS18037A). F.RJ.\ was partly supported by MATS 8.0 program during which foundational experiments for this work were
completed.
\end{ack}

\newpage
\bibliographystyle{unsrtnat}
\bibliography{main}

\newpage
\appendix

\section{Experimental Details}
\label{sec:experimental_details}
\subsection{LRP Implementation}
We applied the LN-rule \cite{DBLP:conf/icml/AliSEMMW22} for LayerNorm/RMSNorm, the Identity-rule \cite{jafari2024mambalrp} for nonlinear activation functions, and the LRP-$0$ rule \cite{montavon2019layer} for linear transformations. The Half-rule \cite{jafari2024mambalrp, Arras2019} was applied only to the Qwen2 model family and Gemma2-2B, as other models used in our experiments did not include multiplicative gating operations. No attention-specific rule (e.g., AH-rule) was used. Full details are given in Table~\ref{tab:lrp_implementation_details}.

\begin{table}[tbh]
\centering
\renewcommand{\arraystretch}{1.2}
\newcommand{\cmark}{\textcolor{ForestGreen}{\checkmark}}
\newcommand{\xmark}{\textcolor{red}{\texttimes}}
\small
\resizebox{0.85\textwidth}{!}{
\begin{tabular}{lcccc}
\toprule
\textbf{Propagation Rule} & \textbf{GPT-2 Family} & \textbf{Pythia Family} & \textbf{Qwen2 Family} & \textbf{Gemma2-2B}\\
\midrule
LN-rule \cite{DBLP:conf/icml/AliSEMMW22} & \cmark & \cmark & \cmark & \cmark \\
Identity-rule \cite{jafari2024mambalrp} & \cmark & \cmark & \cmark & \cmark \\
$0$-rule \cite{montavon2019layer} & \cmark & \cmark & \cmark & \cmark \\
Half-rule \cite{jafari2024mambalrp, Arras2019} & \xmark & \xmark & \cmark & \cmark \\
AH-rule \cite{DBLP:conf/icml/AliSEMMW22} & \xmark & \xmark & \xmark & \xmark \\
\bottomrule
\end{tabular}
}
\vspace{0.5em}
\caption{Propagation rules applied in our LRP implementation across different model families. A green tick (\cmark) indicates that the rule was applied, while a red cross (\xmark) indicates that it was not used.}
\label{tab:lrp_implementation_details}
\end{table}

\subsection{Further Details on Subject-Verb Agreement Experiment}
In this experiment, we used 300 samples, structured according to Table~\ref{tab:other_templates}, for circuit discovery, and 100 held-out samples for evaluating faithfulness and completeness. Consistent with \cite{marks2024sparse}, the first one-third of each circuit was excluded from evaluation, since components in early model layers are typically responsible for processing specific tokens, which may not consistently appear across training and test splits.

\section{Quantitative results}
\label{sec:quantitative_results}
The exact numerical values corresponding to the IOI experiment, visualized in Figure~\ref{fig:quant_results}, are reported in Table~\ref{tab:pcc}.

\begin{table}[ht]
  \centering
  \resizebox{0.8\textwidth}{!}{
  \begin{tabular}{lcccccc}
    \toprule
      & \multicolumn{2}{c}{Residual Stream}
      & \multicolumn{2}{c}{Attention Outputs}
      & \multicolumn{2}{c}{MLP Outputs} \\
    \cmidrule(lr){2-3} \cmidrule(lr){4-5} \cmidrule(lr){6-7}
    Model
      & AtP & RelP (ours)
      & AtP & RelP (ours)
      & AtP & RelP (ours) \\
    \midrule
    GPT-2 small
      &  0.753 & \textbf{0.968}
      &  \textbf{0.979} & 0.962
      &  0.195 & \textbf{0.992} \\
    GPT-2 medium
      &  0.671 & \textbf{0.884}
      &  0.961 & \textbf{0.982}
      &  0.551 & \textbf{0.993} \\
    GPT-2 large
      &  0.740 & \textbf{0.853}
      &  0.461 & \textbf{0.965}
      &  0.006 & \textbf{0.956} \\
    Pythia-70M
      &  0.728 & \textbf{0.898}
      &  0.981 & \textbf{0.986}
      &  0.507 & \textbf{0.967} \\
    Pythia-410M
      &  0.695 & \textbf{0.874}
      &  0.879 & \textbf{0.903}
      &  0.501 & \textbf{0.948} \\
    Qwen2-0.5B
      &  0.684 & \textbf{0.922}
      &  0.933 & \textbf{0.965}
      &  0.717 & \textbf{0.981} \\
    Qwen2-7B
      &  0.230 & \textbf{0.549}
      &  0.395 & \textbf{0.569}
      &  0.471 & \textbf{0.622} \\
    Gemma2-2B
      &  0.391 & \textbf{0.769}
      &  0.942 & \textbf{0.956}
      &  0.910 & \textbf{0.952} \\
    \bottomrule
  \end{tabular}
  }
  \vspace{0.5em}
  \caption{Pearson correlation coefficient (PCC) between activation patching and attribution patching (AtP) or relevance patching (RelP), computed over 100 IOI prompts for three GPT-2 model sizes (Small, Medium, Large), two Pythia models (70M, 410M), two Qwen2 models (0.5B, 7B), and Gemma2-2B. A higher value of PCC represents higher alignment with activation patching results.}
  \label{tab:pcc}
\end{table}

\section{Further Qualitative Results}
\label{sec:further_qualitative_results}
In Section~\ref{sec:ioi}, we presented the qualitative differences between AtP and RelP for the GPT-2 Small model. In this section, we extend our analysis by providing additional experimental results for other models. It is evident that RelP provides a more accurate approximation to activation patching, particularly for the residual stream and MLP outputs. 

\begin{figure}
    \centering
    \includegraphics[width=\linewidth]{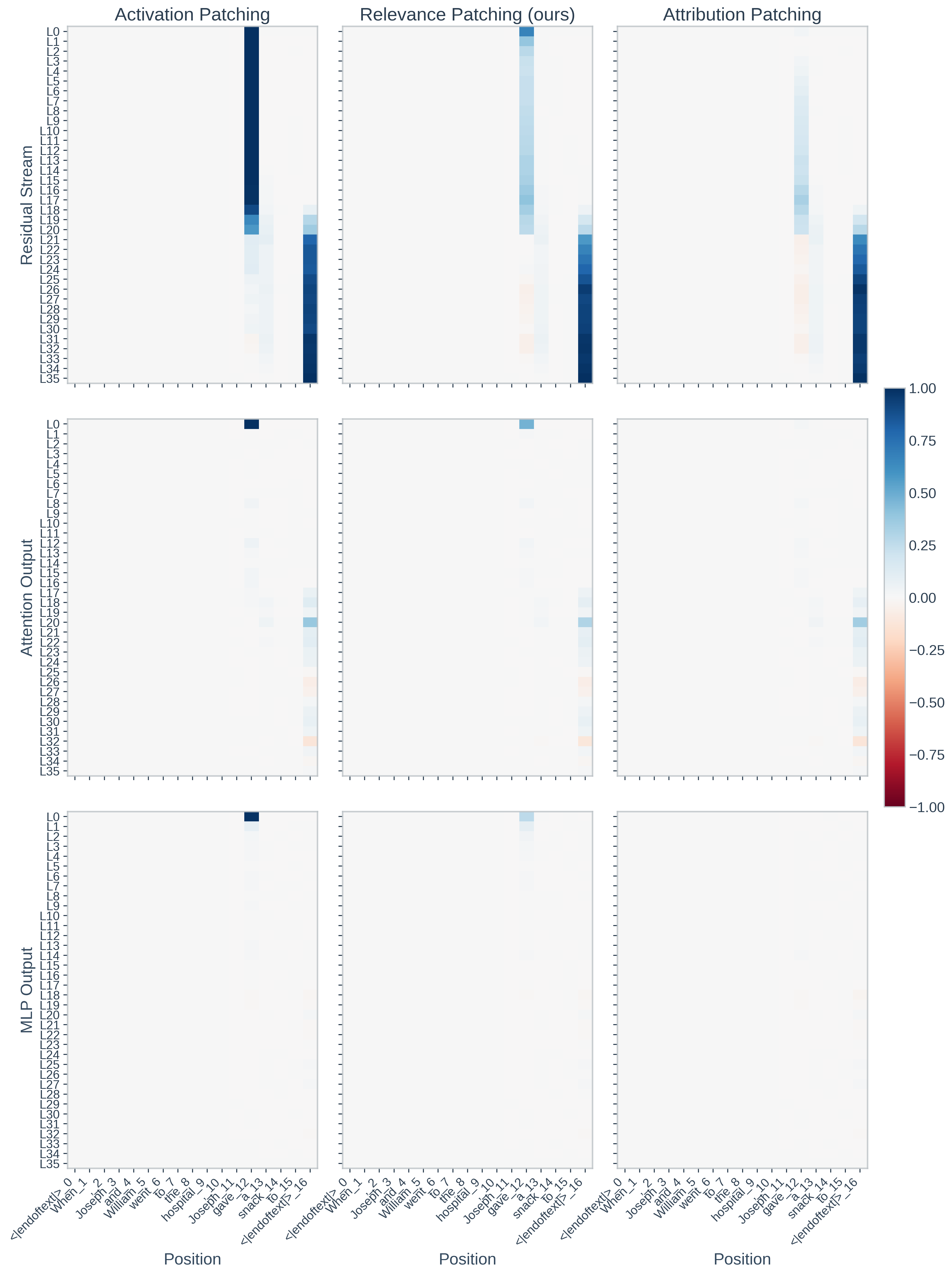}
    \caption{Qualitative comparison showing how accurately relevance patching (RelP) and attribution patching (AtP) approximate the effects of activation patching in GPT-2 Large. RelP shows notably better alignment in the residual stream and at MLP0, where AtP's estimates are less reliable.}
    \label{fig:ioi_qualitative_gpt2_large}
\end{figure}

\begin{figure}
    \centering
    \includegraphics[width=\linewidth]{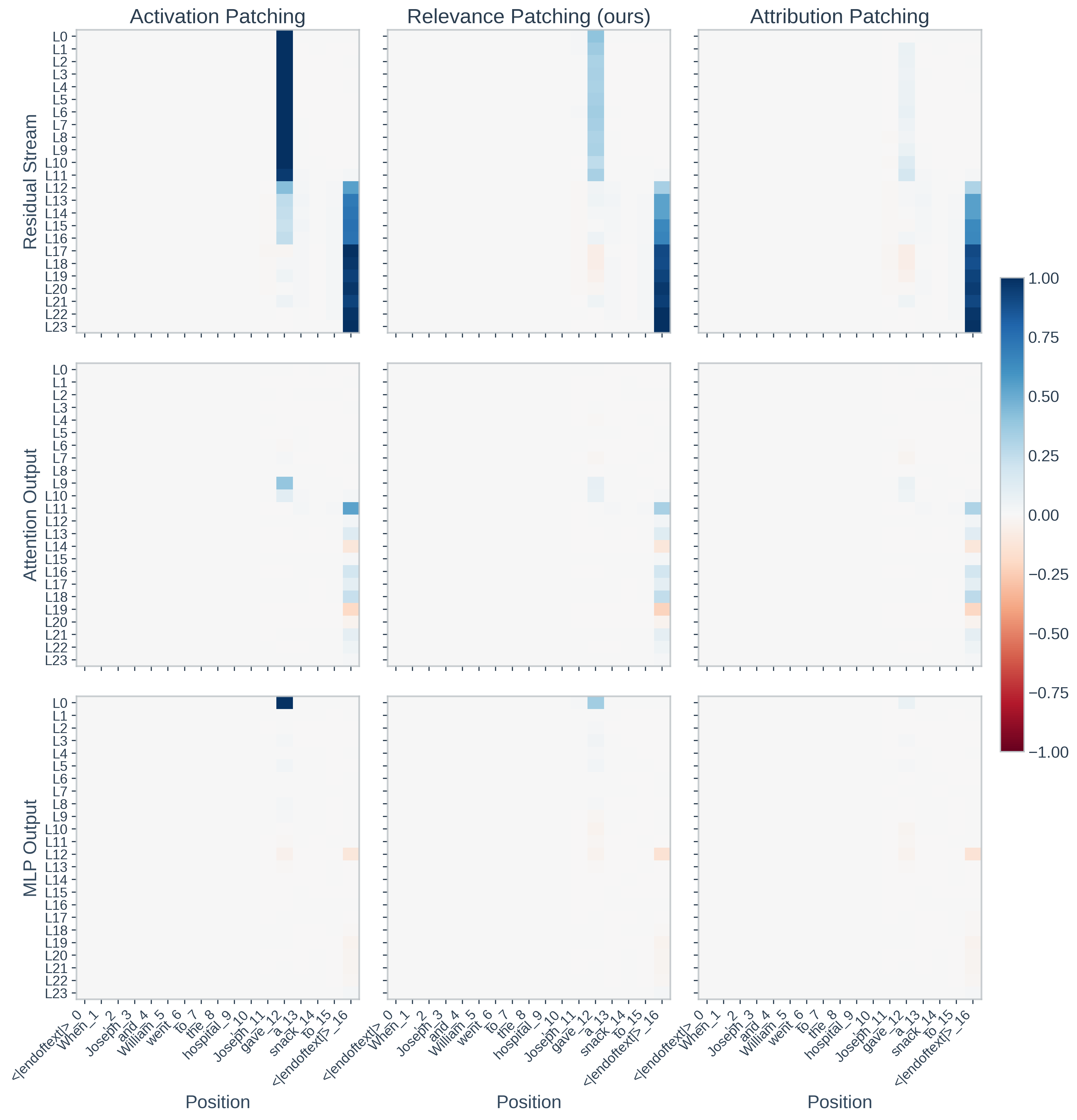}
    \caption{Qualitative comparison showing how accurately relevance patching (RelP) and attribution patching (AtP) approximate the effects of activation patching in Pythia-410M. RelP shows notably better alignment in the residual stream and at MLP0, where AtP's estimates are less reliable.}
    \label{fig:ioi_qualitative_pythia_410}
\end{figure}

\end{document}